\documentclass[review]{elsarticle}

\usepackage{lineno,hyperref}

\usepackage{amsthm,amsmath,amssymb}
\usepackage{mathrsfs}
\usepackage{multirow}
\usepackage{array}
\usepackage{graphicx}

\usepackage{multirow}
\usepackage{float}
\usepackage{booktabs}
\usepackage{xcolor}
\usepackage{url}
\usepackage{color}
\usepackage{array}
\usepackage{tikz}
\usepackage{hyperref}
\usepackage{colortbl}
\usepackage[T1]{fontenc}

\journal{xxx}

\begin{document}

\begin{frontmatter}



\title{Deep Self-cleansing for Medical Image Segmentation with Noisy Labels}

\author{Jiahua Dong$^{a}$\fnref{equal}}\author{Yue Zhang$^{b,c}$\fnref{equal}}\author{Qiuli Wang$^{d}$}\author{Ruofeng Tong$^{a}$\corref{mycorrespondingauthor}}\author{Shihong Ying$^{e}$\corref{mycorrespondingauthor}}\author{Shaolin Gong$^{e}$}\author{Xuanpu Zhang$^{e}$}\author{Lanfen Lin$^{a}$}\author{Yen-Wei Chen$^{f}$} \author{S. Kevin Zhou$^{b,c,g,i}$\corref{mycorrespondingauthor}}
\address{a College of Computer Science and Technology, Zhejiang University, Hangzhou, China}
\address{b School of Biomedical Engineering, Division of Life Sciences and Medicine,
University of Science and Technology of China, Hefei, Anhui, China}
\address{c Center for Medical Imaging, Robotics, Analytic
Computing \& Learning (MIRACLE),
Suzhou Institute for Advanced Research, University of Science and Technology
of China, Suzhou, Jiangsu, China}
\address{d 7T Magnetic Resonance Translational Medicine Research Center, Department of Radiology, Southwest Hospital, Army Medical University (Third Military Medical University), Chongqing, China}
\address{e Department  of  Radiology, The  First Affiliated  Hospital,  College  of  Medicine,  Zhejiang  University,  Hangzhou, China}
\address{f College of Information Science and Engineering, Ritsumeikan University, Shiga, Japan}
\address{g Key Laboratory of Precision and Intelligent Chemistry, University of Science and Technology of China, Hefei
Anhui, China}
\address{i Key Laboratory of Intelligent Information Processing of Chinese Academy of Sciences (CAS), Institute of Computing Technology, CAS}

\fntext[equal]{Jiahua Dong and Yue Zhang contribute equally to this work.}

\begin{abstract}
Medical image segmentation is crucial in the field of medical imaging, aiding in disease diagnosis and surgical planning. Most established segmentation methods rely on supervised deep learning, in which clean and precise labels are essential for supervision and significantly impact the performance of models. However, manually delineated labels often contain noise, such as missing labels and inaccurate boundary delineation, which can hinder networks from correctly modeling target characteristics.
In this paper, we propose a deep self-cleansing segmentation framework that can preserve clean labels while cleansing noisy ones in the training phase. To achieve this, we devise a gaussian mixture model-based label filtering module that distinguishes noisy labels from clean labels. Additionally, we develop a label cleansing module to generate pseudo low-noise labels for identified noisy samples. The preserved clean labels and pseudo-labels are then used jointly to supervise the network.
Validated on a clinical liver tumor dataset and a public cardiac diagnosis dataset, our method can effectively suppress the interference from noisy labels and achieve prominent segmentation performance.
\end{abstract}




\end{frontmatter}




\section{Introduction}
\label{Intro}
Medical image segmentation is a fundamental task in medical image analysis, which delineates specific organs or tumors from medical images such as computed tomography (CT) and Magnetic Resonance Imaging (MRI). Accurate segmentation provides critical diagnostic information, including shapes, locations, and textures of target tissues, thereby assisting clinicians in disease diagnosis and surgical planning~\cite{zheng2022automatic,di2022td}.
Recently, supervised deep learning-based methods~\cite{ronneberger2015u,zhou2018unet++,oktay2018attention,seo2019modified,huang2020unet,xiao2018weighted,li2018h,chen2021transunet,hatamizadeh2022unetr,jiang2022swinbts,lin2022ds,cao2022swin,cciccek20163d,milletari2016v} have achieved immense success in medical image segmentation, with U-Net~\cite{ronneberger2015u} being the most popular one. Based on U-Net, a lot of variants have been developed by improving skip connections~\cite{zhou2018unet++,oktay2018attention,seo2019modified,huang2020unet}, embracing transformers~\cite{chen2021transunet,hatamizadeh2022unetr,jiang2022swinbts,lin2022ds,cao2022swin}, etc.
\begin{figure}[t]
\centering
\includegraphics[width=0.7\textwidth]{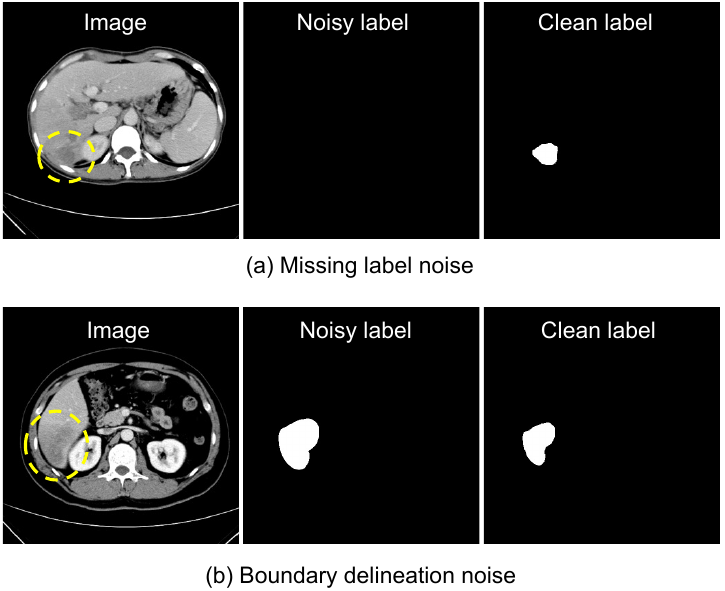}
\caption{Typical types of label noise. Take CT slices with liver tumors as an example, (a) illustrates a missing label that radiologists omitted the tumor region. (b) illustrates the boundary noises around the tumor region.
}
\label{fig_lb_error}
\end{figure}
Supervised methods generally rely on high-quality, large-scale training datasets annotated with clean labels that ensure networks learn correct representations. However, in clinical practice, obtaining flawless labels is hardly possible~\cite{zhu2019pick} since radiologists inevitably experience visual fatigue, which leads to inconsistent and noisy labels. 

Typically, label noise within medical images can be grouped into missing label noise (as shown in Fig.~\ref{fig_lb_error}(a)) and boundary delineation noise (as shown in Fig.~\ref{fig_lb_error}(b)). Missing label noise occurs when radiologists omit target-included slices, resulting in target tissues that are not delineated. Boundary delineation noise, on the other hand, occurs when the low contrast between target tissues and surroundings makes it difficult for radiologists to accurately determine the boundaries, hence leading to inaccurate edge delineations.
Training datasets corrupted by label noise can degrade the performance of deep models, as the models learn incorrect characteristics of targets. Therefore, it is crucial to develop methods that make networks resistant to such noise.

In recent years, learning with noisy labels has attracted considerable attention. Most existing methods focus on classification tasks with natural images, including but not limited to designing robust losses~\cite{cheng2020learning,ghosh2017robust,zhang2018generalized}, re-weighting samples~\cite{liu2015classification,ren2018learning,shu2019meta}, and label correction~\cite{han2019deep,tanaka2018joint,zheng2020error}. However, learning segmentation with noisy labels in medical images has not been extensively investigated. Few known methods in this field can be classified into two categories: \textbf{1)} Image-level cleansing methods, which estimate the overall noise level of each image and reduce the impact of noisy samples on network training~\cite{zhu2019pick}. 
For example, Zhu et al.~\cite{zhu2019pick} proposed a quality awareness module to evaluate the quality of labels in the training set, and assign lower weights for noisy labels when constructing losses for chest image segmentation.
\textbf{2)} Pixel-level cleansing methods, which identify specific noisy regions (pixels) within each image and modify the labels through pseudo-labeling~\cite{liu2021s,zhang2020robust,zhang2020characterizing,xu2021noisy,wei2020combating, li2021superpixel}. 
For example, Liu et al.~\cite{liu2021s} and Zhang et al.~\cite{zhang2020robust} used network predictions as corrected labels to cleanse noisy labels. Zhang et al.~\cite{zhang2020characterizing} and Xu et al.~\cite{xu2021noisy} employed Confident Learning~\cite{northcutt2021confident} to estimate wrongly-labeled pixels and make corrections. Wei et al.~\cite{wei2020combating} used two different networks to select confident pixels within labels for cross-training. Li et al.~\cite{li2021superpixel} integrated superpixel representations to guide the label refinement process.
Although image-level methods could suppress the influence of noisy samples, local noisy regions within low-noise samples still affect the performance; Meanwhile, pixel-level methods could not distinguish between noisy and clean samples, leading to the indiscriminate modification of all labels (including clean and low-noise labels), which is inefficient and prone to introducing new noise. 

To address the above issues, we combine the merits of image-level and pixel-level cleansing methods for medical image segmentation with noisy labels. We propose a deep self-cleansing network that preserves clean labels while cleansing noisy labels in an iterative manner. First, we devise an image-level label filtering module (LFM) to distinguish between noisy and clean labels. Given that clean labels typically generate smaller loss values than noisy ones in the early training stage~\cite{chen2019understanding}, the LFM uses Gaussian Mixture Models (GMM)~\cite{reynolds2009gaussian} to model loss distributions and classify labels into \emph{Clean} (need preserving) and \emph{Noisy} (need cleansing) categories. To cleanse the \emph{Noisy} labels, we then propose a pixel-level label cleansing module (LCM) based on pseudo-labeling. The LCM extracts representative target regions from network outputs as prototypes, and computes pseudo-labels based on the similarity between pixel positions and prototypes.
Finally, the preserved \emph{Clean} labels and generated pseudo-labels are used to jointly supervise the network. Extensive experiments on an abdominal CT dataset of liver tumors and an MRI dataset for cardiac diagnosis demonstrate that the proposed self-cleansing framework is resistant to label noise and achieves excellent segmentation performance.

In summary, our work makes the following contributions:

\begin{itemize}
\item We present a noise-resistant framework for medical image segmentation with noisy labels. The devised framework can preserve clean labels and iteratively cleanse noisy labels during the training phase.

\begin{table}[t]
\centering
\footnotesize
\caption{Detailed structure of the segmentation backbone.}
\label{tb_backbone}
\begin{tabular}{lll} 
\hline\hline
\multirow{2}{*}{\textbf{Layer Name}} & \multirow{2}{*}{\textbf{Output Size}} & \textbf{Convolution, kernel size},~  \\
                                     &                                       & \textbf{output channels, stride}    \\ 
\hline
Input                               & 256×256                               & -                                    \\\hline
Conv\_1                              & 256×256                               & Conv, 3×3, 16, 1                     \\\hline
\multirow{2}{*}{Residual\_block\_1}  & \multirow{2}{*}{128×128}              & Conv, 2×2, 32, 2                     \\
                                     &                                       & Conv, 3×3, 32, 1                     \\\hline
\multirow{3}{*}{Residual\_block\_2}  & \multirow{3}{*}{64×64}                & Conv, 2×2, 64, 2                     \\
                                     &                                       & Conv, 3×3, 64, 1                     \\
                                     &                                       & Conv, 3×3, 64, 1                     \\\hline
\multirow{4}{*}{Residual\_block\_3}  & \multirow{4}{*}{32×32}                & Conv, 2×2, 128, 2                    \\
                                     &                                       & Conv, 3×3, 128, 1                    \\
                                     &                                       & Conv, 3×3, 128, 1                    \\
                                     &                                       & Conv, 3×3, 128, 1                    \\\hline
\multirow{4}{*}{Residual\_block\_4}  & \multirow{4}{*}{64×64}                & ConvTranspose, 2×2, 64, 2            \\
                                     &                                       & Conv, 3×3, 128, 1                    \\
                                     &                                       & Conv, 3×3, 128, 1                    \\
                                     &                                       & Conv, 3×3, 128, 1                    \\\hline
\multirow{3}{*}{Residual\_block\_5}  & \multirow{3}{*}{128×128}              & ConvTranspose, 2×2, 32, 2            \\
                                     &                                       & Conv, 3×3, 64, 1                     \\
                                     &                                       & Conv, 3×3, 64, 1                     \\\hline
\multirow{2}{*}{Residual\_block\_6}  & \multirow{2}{*}{256×256}              & ConvTranspose, 2×2, 16, 2            \\
                                     &                                       & Conv, 3×3, 32, 1                     \\\hline
\multirow{3}{*}{Conv\_2}             & \multirow{3}{*}{256×256}              & Conv, 3×3, 2, 1                      \\
                                     &                                       & Conv, 1×1, 1, 1                      \\
                                     &                                       & Sigmoid                              \\
\hline\hline
\end{tabular}
\end{table}
\item We present a GMM-based image-level label filtering module to distinguish between noisy and clean labels. Based on this module, our framework focuses on refining only noisy labels, thereby making the training stage stable and controllable.
\item We devise a pixel-level label cleansing module to correct noisy labels by identifying representative prototypes and generating low-noise pseudo-labels.
\item We verify the proposed method on an abdominal CT dataset of liver tumors and a public MRI dataset for cardiac diagnosis. Experimental results demonstrate the effectiveness and robustness of our method when training with noisy labels.

\end{itemize}

\begin{figure*}[t]
\centering
\includegraphics[width=\textwidth]{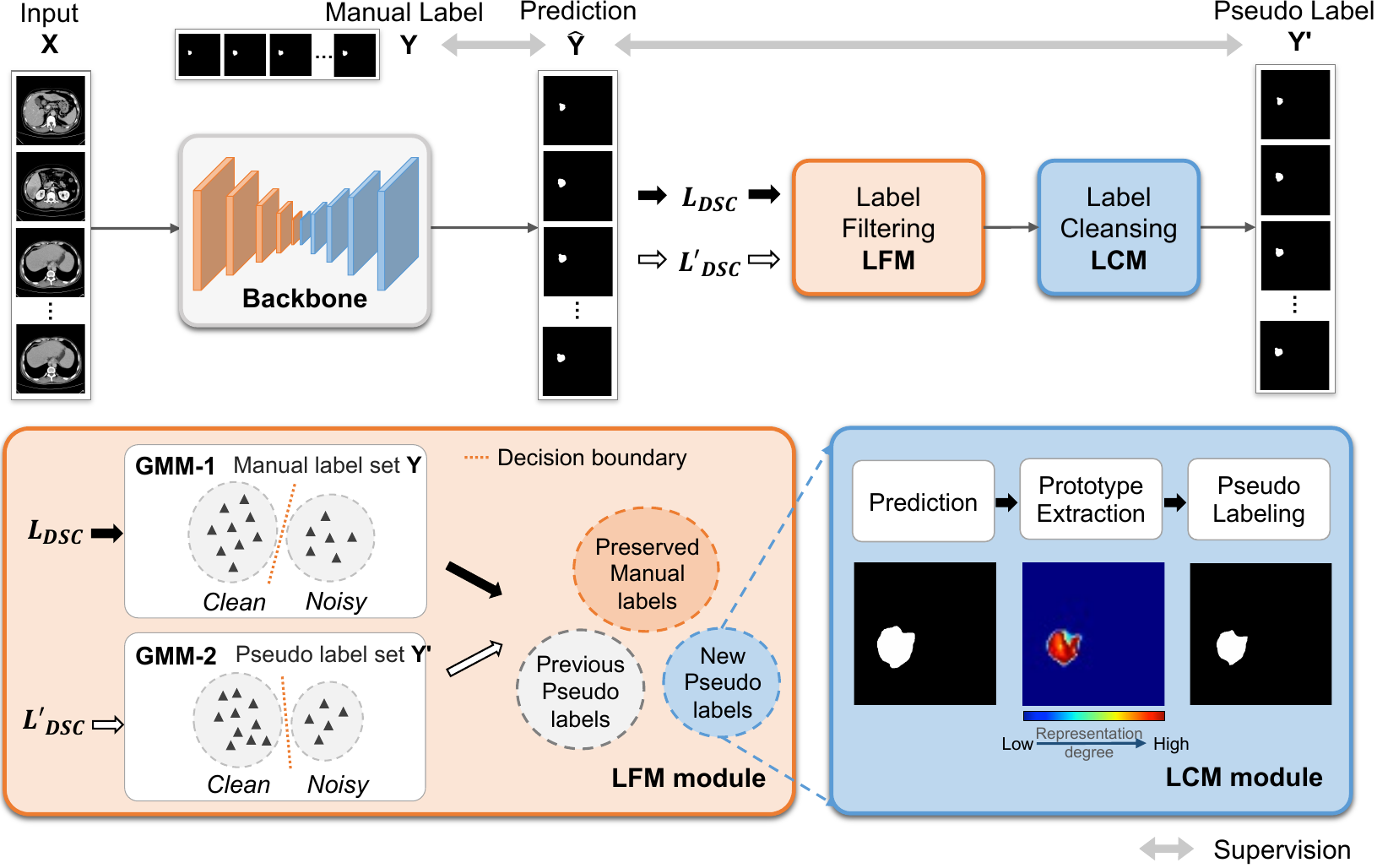}
\caption{Schematic view of the deep self-cleansing network, which filters out the \emph{Noisy} labels through LFM and cleanses them through LCM interatively.}
\label{fig_overview}
\end{figure*}

\section{Method}
\label{method}

Fig.~\ref{fig_overview} illustrates the schematic view of the proposed self-cleansing method for medical image segmentation. This method comprises three main components: the segmentation backbone, the label filtering module (LFM), and the label cleansing module (LCM). During training, our network iteratively preserves clean labels and cleanses noisy ones using LFM and LCM. In each cleansing iteration, the LFM first distinguishes noisy labels from clean labels based on per-sample loss distributions. Subsequently, the LCM generates low-noise pseudo-labels for the identified noisy samples based on class prototypes. The two modules are applied every five epochs throughout the training phase. The following subsections will provide detailed descriptions of each component.

\subsection{Segmentation Backbone}
\label{sec_backbone}
The proposed framework can use existing segmentation networks as its backbone. Given the U-Net architecture's effectiveness in combining high- and low-level features for medical image segmentation~\cite{ronneberger2015u}, we employ an enhanced version of U-Net with residual connections as the segmentation backbone in this paper. Our backbone comprises three down-sampling stages and three up-sampling stages to extract multi-level and multi-scale features progressively. The specific network structure is detailed in Table~\ref{tb_backbone}.
To ensure that the backbone network can extract effective features and be prepared for subsequent noisy label cleansing, we initially warm up the backbone network for 50 epochs.

\subsection{Image-Level Label Filtering}
\label{image_level}
Existing pixel-level cleansing methods generate pseudo-labels for all samples. However, in real-world scenarios, annotated samples often contain varying levels of label noise. Modifying all labels indiscriminately can introduce new noise into clean and low-noise labels, thereby impacting the model's stability and performance. Therefore, the first step in label self-cleansing is to filter out noisy labels for cleansing while preserving the clean labels.

\textbf{Gaussian Mixture Model for Label Classification:}
It has been observed that neural networks tend to first fit clean labels during the early training stage~\cite{arpit2017closer, chen2019understanding}. During this period, the network's predictions usually result in smaller losses for clean labels, whereas the losses for noisy labels are typically larger. 
Therefore, our LFM leverages the loss characteristic to filter out noisy samples.
Specifically, we use the Dice Similarity Coefficient (DSC) loss~\cite{milletari2016v}, which is the most commonly used metric in segmentation, as an example to illustrate this process.

Let $X={\left\{ x_i \right\}}_{i=1}^N$ denote the input images (where $N$ is the number of training samples), $Y={\left\{ y_i \right\}}_{i=1}^N$ denote the manual labels, and $\hat{Y}={\left\{ \hat{y}_i \right\}}_{i=1}^N$ denote the network predictions, the DSC loss between $\hat{Y}$ and $Y$ is formulated as:
\begin{equation}
    L_{DSC}={\{{l_i \}}}_{i=1}^N=\{1-\frac{2\left | \hat{y_{i}}\cap y_{i} \right |}{\left | \hat{y_{i}} \right |+\left | y_{i} \right |} \}_{i=1}^{N}
\end{equation}

Given the flexibility of the Gaussian Mixture Model (GMM) in fitting distributions with varying sharpness~\cite{li2020dividemix}, we employ a two-component GMM to model per-sample loss distributions and to classify the training labels into \emph{Clean} (to be preserved) and \emph{Noisy} (to be cleansed) categories.
After normalizing $L_{DSC}$ into the range of [0,1], we use the Exception Maximization algorithm~\cite{moon1996expectation} to fit a two-component GMM to the distribution of $L_{DSC}$. This GMM allows us to compute the posterior probabilities of each label $y_i$ being classified as either \emph{Clean} or \emph{Noisy}. These posterior probabilities are denoted as $w_{Clean}^{(i)}$ and $w_{Noisy}^{(i)}$, respectively:
\begin{equation}
    w_{Clean}^{(i)}=p(g_{small}|l_i)
\end{equation}
\begin{equation}
    w_{Noisy}^{(i)}=p(g_{large}|l_i)
\end{equation}
where $g_{small}$ is the gaussian component with the smaller mean value, and $g_{large}$ is the gaussian component with the larger mean value. We classify each label $y_i$ as \emph{Clean} if $w_{Clean}^{(i)}>w_{Noisy}^{(i)}$; otherwise, it is classified as \emph{Noisy}.

\textbf{Cascaded Label Filtering:}
In this paper, our network is devised to filter and cleanse noisy labels through pseudo-labeling every $k$ epochs, making the training stage an iterative cleansing process. To ensure efficient label cleansing and minimize potential noise introduced by pseudo-labeling, we retain both \emph{Clean} manual labels and \emph{Clean} previous pseudo-labels (generated in the previous iteration) and modify only the remaining labels. To achieve this, we propose a cascaded label filtering algorithm using two GMMs to filter the labels that require cleansing.

Let $Y'={\left\{ y'_i \right\}}_{i=1}^N$ denote the pseudo label set, initially containing all-zero images, and let $L_{DSC}'={\{{l'_i \}}}_{i=1}^N$ denote losses between $\hat{Y}$ and $Y'$. We first fit a GMM (termed \textbf{GMM-1} in Fig.~\ref{fig_overview}) to $L_{DSC}$ to classify $Y$ into \emph{Noisy} and \emph{Clean} categories. Subsequently, we fit another GMM (termed \textbf{GMM-2} in Fig.~\ref{fig_overview}) to $L'_{DSC}$ to categorize $Y'$ into \emph{Noisy} and \emph{Clean} categories. Finally, we cleanse only those samples whose both manual label and pseudo label are identified as \emph{Noisy}. The new filtering criteria is as follows: 
\begin{itemize}
\item[-] If a sample $x_i$ has a \emph{Clean} manual label $y_{i}$, its pseudo label $y'_{i}$ is kept the same as $y_{i}$, thereby preserving the \emph{Clean} manual label;
\item[-] If a sample $x_i$ has a \emph{Noisy} manual label and a \emph{Clean} pseudo label, $y'_{i}$ remains unchanged, thus preserving \emph{Clean} previous pseudo-label;
\item[-] If both the manual label and pseudo label of $x_i$ are \emph{Noisy}, the sample should be filtered out. And its corresponding $y'_{i}$ will be cleansed by the LCM with newly generated pseudo labels.
\end{itemize}

After each cleansing iteration, $Y'$ consists of three parts as shown in Fig.~\ref{fig_overview}: \textbf{1)} preserved \emph{Clean} manual labels, \textbf{2)} preserved \emph{Clean} pseudo labels generated in the previous iteration, and \textbf{3)} newly generated pseudo-labels. By maintaining a portion of low-noise manual labels, $Y'$ ensures the stability of the training process and mitigates cumulative errors to a certain extent.

\begin{figure}[t]
\centering
\includegraphics[width=\textwidth]{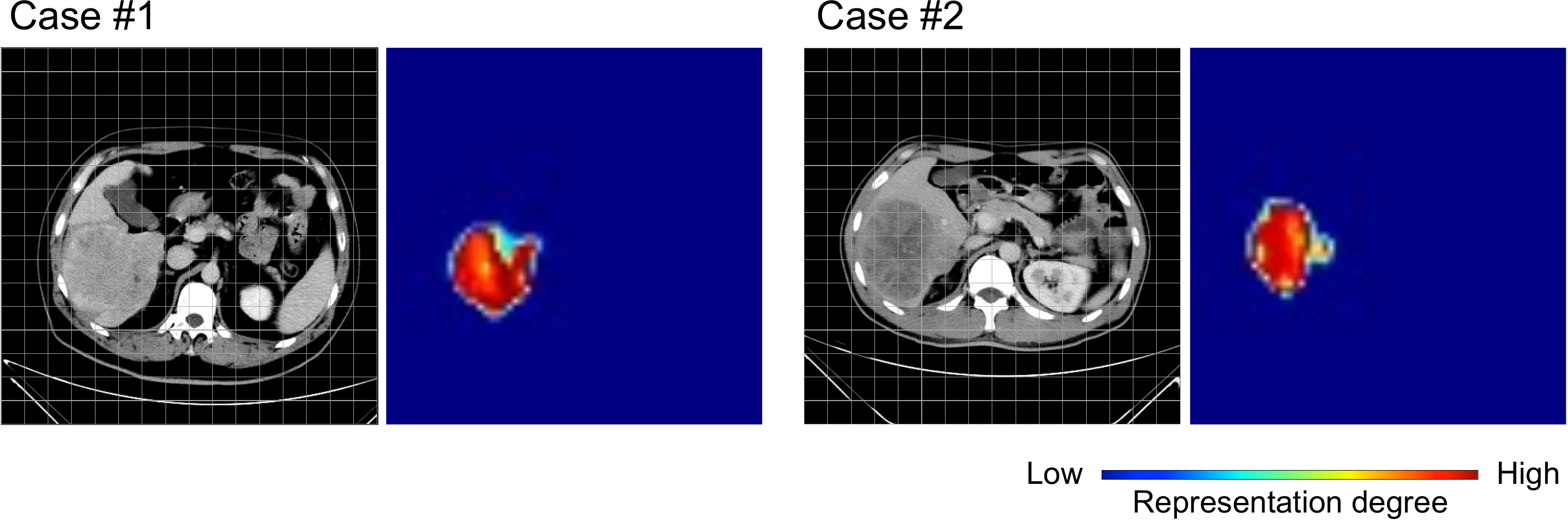}
\caption{Illustration of grid-based representation maps.}
\label{fig_grid}
\end{figure}

\subsection{Pixel-Level Label Cleansing}
\label{sec_lcm}
Sec.~\ref{image_level} introduces how LFM preserves clean labels and filters out noisy labels to be cleansed. Here, we elaborate on the details of LCM  based on pseudo-labeling. Known pseudo-labeling methods typically measure feature similarity between pixels and target prototypes to compute new classification scores. Prototypes are often calculated as either the average feature of the suspected region (network predictions) or as several feature vectors of discrete representative pixels. However, these methods struggle to accurately represent local characteristics of targets, leading to inaccuracies in pseudo-label generation. To address this issue, LCM refines the process by selecting the most representative region from the suspected area as prototypes, while discarding low-confidence regions. The features from this refined representative region are then used as prototype features for calculating pseudo-labels.

\begin{figure*}[t]
\centering
\includegraphics[width=\textwidth]{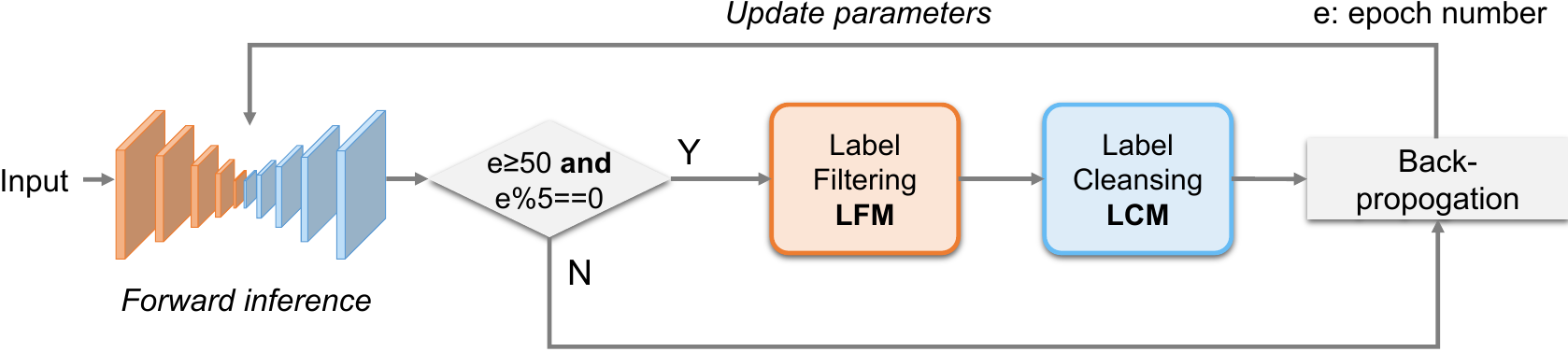}
\caption{Illustration of the training pipeline.} \label{fig_scheme}
\end{figure*}

\textbf{Prototype Construction:}
For an input image $x_i$ (abbreviated as $x$), its feature map, i.e., the output of "Residual\_block\_6" in Table~\ref{tb_backbone}, is denoted as $f$, and the suspected target region is denoted as $R$. We construct prototypes based on representative features from $f$. Specifically, we divide $f$ into $K \times K$ grids ($K=64$ in experiments) and select the grids most representative of $R$. The features of these representative grids are then used as prototypes. Note that we process feature maps based on grids instead of pixels, as calculating the representation degree for each pixel sequentially is highly memory-intensive.

Let $G={\{g_j\}}_{j=1}^{K^2}$ denote the grids, the feature of a grid $g_j$ is an aggregated feature calculated as follows:
\begin{equation}
    f(g_j)=\frac{\sum_{p\in g_j}w(p)f(p)}{\sum_{p\in g_j}w(p)},\quad w(p)=\left\{\begin{matrix}
1, & p\in R\\ 
0, & p\notin R
\end{matrix}\right.
\label{eq_grid_feature}
\end{equation}
where $p$ is a pixel position in $g_j$ and $f(p)$ is the feature vector of $p$. The representation degree $D$ of each grid can be measured by the following equation:
\begin{equation}
    D(g_j)=\left\{\begin{matrix}
  \sum_{g_k\in R} \cos \left(f(g_j), f(g_k)\right), & g_j\in R\\ 
  0, & g_j\notin R
\end{matrix}\right.
\end{equation}
where $\cos \ (*,*)$ denotes the cosine similarity. $D(g_j)$ reflects the similarity between $g_j$ and $R$. A larger $D(g_j)$ indicates that $g_j$ is more representative of target areas. Fig.~\ref{fig_grid} provides examples of grid-based representation maps.
After normalizing $D$ to the range [0,1], we set a threshold $\sigma $ (empirically set to 0.7) to extract grids with $D>\sigma$ as representative grids. Finally, the features of these representative grids, calculated using Eq.~\ref{eq_grid_feature}, are used as prototypes, denoted as $PROTO={\{proto_m\}}_{m=1}^M (M<K^2)$.

\textbf{Pseudo Labeling:}
Subsequently, LCM generates a pseudo label for $x$ by updating the classification score of each pixel. For a pixel $q\in x$, its new foreground probability $P_{fg}(q)$ is calculated by measuring feature similarity between $q$ and prototypes:
\begin{equation}
    P_{fg}(q)=\frac{1}{M}\sum_{m=1}^{M}\cos (f(q),proto_m)
\end{equation}
The final pseudo label can be obtained by converting the soft probability map into a binary mask with a threshold $\mu$ (empirically set to 0.7).

After obtaining the pseudo-labels, we further refine them through post-processing, which involves two steps: 1) Removing connected foreground regions with small areas. Since the generated pseudo-labels may contain fragmented regions, we empirically remove any connected foreground regions whose area is smaller than 10\% of the total foreground area; 2) Filling small holes within the foreground. Similarly, we fill any holes with an area smaller than 10\% of the total foreground area.

\subsection{Training Scheme}
\label{sec_loss}
\textbf{Loss Function:}
We devise a hybrid loss $L_{\text{hybrid}}$ to optimize the proposed network, which can be expressed as:
\begin{equation}
    L_{\text{hybrid }}=\lambda_1(e) L_{DSC}+\lambda_2(e) L'_{DSC}
\end{equation}
where $L_{DSC}$ measures the difference between network outputs and manual labels, while $L'_{DSC}$ reflects the difference between outputs and pseudo labels. The coefficients $\lambda_1(e)$ and $\lambda_2(e)$ are two epoch-dependent parameters, which are defined as:
\begin{equation}
\begin{split}
    \lambda_1(e)=\left\{\begin{array}{ll}
1, & 0 \leq e<E_{1} \\
\frac{E_{2}-e}{E_{2}-E_{1}}, & E_{1} \leq e<E_{2} \\
0, & E_{2} \leq e<E
\end{array}\right.
\\
\lambda_2(e)=\left\{\begin{array}{ll}
0, & 0 \leq e<E_{1} \\
\frac{e-E_{1}}{E_{2}-E_{1}}, & E_{1} \leq e<E_{2} \\
1, & E_{2} \leq e<E
\end{array}\right.
\end{split}
\end{equation}
where $e$ represents the training epoch, $E$ is the total training epochs ($E=500$), $E_1$ is the warm-up epochs ($E_1=E/10$), and $E_2=E/2$.

\textbf{Training Pipeline:}
The entire training pipeline is illustrated in Fig.~\ref{fig_scheme}.
In the early training stage, our network relies only on manual labels to learn features. After a warm-up period of 50 epochs, the network begins to update pseudo-labels every $k$ epochs (set to 5 in our experiments), with the contribution of pseudo-labels gradually increasing. As the training becomes stable, the network is fine-tuned using only pseudo-label set. It is important to note that the pseudo-label set always includes some \emph{Clean} manual labels to ensure that the training process remains controlled.

Our method is implemented by PyTorch 1.5.0~\cite{paszke2019pytorch} and deployed on an NVIDIA GTX 2080 Ti GPU (12 GB). We use the SGD optimizer~\cite{bottou2012stochastic} with a mini-batch size of 16. The initial learning rate is set to $4 \times 10^{-4}$ and is divided by 10 every 50 epochs.

\section{Materials and Experiments}
\subsection{Materials}
Two datasets, an in-house CT dataset of liver tumors (HCC dataset) and a public MRI dataset for automatic cardiac diagnosis (ACDC dataset), are used to verify the effectiveness of the proposed method.

The \textbf{HCC dataset}~\cite{zhang2021deeprecs} is an in-house liver tumor dataset collected from the First Affiliated Hospital of Zhejiang University School of Medicine, consisting of 231 cases of CT images (30,828 CT slices in total) of hepatocellular carcinoma. Each case has an intra-slice resolution of $512\times512$, a slice thickness of 0.5 mm, and an inter-plane resolution varying from $\rm 0.56\times 0.56 mm^2$ to $\rm 0.85\times 0.85 mm^2$. The tumor regions within each sample in the dataset is delineated by experienced radiologists. In our study, all slices are resized to $256\times256$ and truncated into the range of [-70, 180] Hu to eliminate irrelevant tissues. The dataset is randomly divided into three parts: 139
CT scans for training, 46 CT scans for validation, and 46 CT scans for testing.

The \textbf{ACDC dataset}~\cite{bernard2018deep} is a publicly available dataset for automated cardiac diagnosis, acquired at the University Hospital of Dijon (France). It consists of 200 MRI scans from 100 patients, obtained using two MRI scanners with magnetic field strengths of 
1.5T and 3.0T.
The scans have in-plane resolutions ranging from $\rm 0.70\times0.70\ mm^2$ to $\rm 1.92\times1.92\ mm^2$, and through-plane resolutions ranging from 5 mm to 10 mm.
All data were collected at the End-Diastolic and End-Systolic stages, with annotations provided for the left ventricle (LV), right ventricle (RV), and myocardium (Myo). 
Following ~\cite{luo2021efficient}, we extracted $128\times128$ patches centered on cardiac regions for our experiments. These patches were further split into 1488 patches for training, 414 patches for validation and 414 patches for testing.

\subsection{Noise Simulation}
To comprehensively evaluate our method under disturbance from varying levels of label noise, we augment the training set by adding label noise of different proportions and amplitudes. 

For the HCC dataset, we randomly select $\alpha_1$ ($\alpha_1 \in \{30\%,50\%,70\%\}$) proportion of images from the training set and randomly add missing-label noise or boundary noises. Specifically, missing-label noise is introduced by removing annotations, while boundary noise is added by randomly dilating or eroding tumor masks (labels) by $\beta_1 \pm 3$ ($\beta_1 \in \{5,10\}$) pixels. This augmentation process results in seven different training sets (including the original set), each corrupted by varying levels of label noise.

Likewise, we generate multiple training sets with different levels of label noise for the ACDC dataset. We randomly select $\alpha_2$ ($\alpha_2 \in {30\%, 50\%, 70\%}$) of the images from the training set and introduce either missing-label noise or boundary noise. Missing-label noise is added by removing the organ masks, while boundary noise is applied by randomly dilating or eroding the organ masks by $\beta_2 \pm 3$ pixels ($\beta_2 \in \{5, 10\}$).

\subsection{Experimental Setup} 
\textbf{Competing Methods:} To evaluate the effectiveness of the proposed method, we first compare it with the \textbf{Baseline} method, i.e., the backbone network without any noise reduction strategy. Besides, we compare our method with four relevant noise-resistance segmentation methods: 
\begin{itemize}
    \item \textbf{Pick-and-learn (PL)}~\cite{zhu2019pick}, an image-level cleansing method that assesses image-level quality of labels in a training batch and assigns lower weights for noisy labels when computing losses.
    \item \textbf{Confident-learning (CL)}~\cite{zhang2020characterizing}, a pixel-level cleansing method that identifies noisy regions within labels and generates pseudo-labels using Confident Learning~\cite{northcutt2021confident}. 
    \item \textbf{Joint Co-Regularization (JoCoR)}~\cite{wei2020combating},a pixel-level cleansing method  that uses two networks to select confident pixels (regions) for cross-training.
    \item \textbf{Superpixel-guided iterative learning (SP)}~\cite{li2021superpixel}, a pixel-level cleansing method that uses superpixel representations to guide noisy label correction.
\end{itemize}

To ensure the fairness of the comparison, all methods adopt the same segmentation backbone (as described in Sec.~\ref{sec_backbone}) and are trained by 500 epochs. Besides, to thoroughly assess the noise-resistance capability of each method, we evaluate the performance of both the B-model and L-model for each method. The B-model represents the best checkpoint selected using the validation set, reflecting the actual segmentation performance on the testing set. Besides, the L-model corresponds to the final checkpoint obtained at the $500^{th}$ epoch, reflecting the stability and sustained noise-resistance performance over time.

\textbf{Evalutation Metrics:} Following PL~\cite{zhu2019pick} and CL~\cite{zhang2020characterizing}, we use the Dice Similarity Coefficient (DSC) metric to quantify segmentation performance by measuring the similarity between the network outputs and the ground truth labels. A higher DSC score indicates better segmentation results and greater robustness to label noise.

\begin{figure}[p]
\centering
\includegraphics[width=\textwidth]{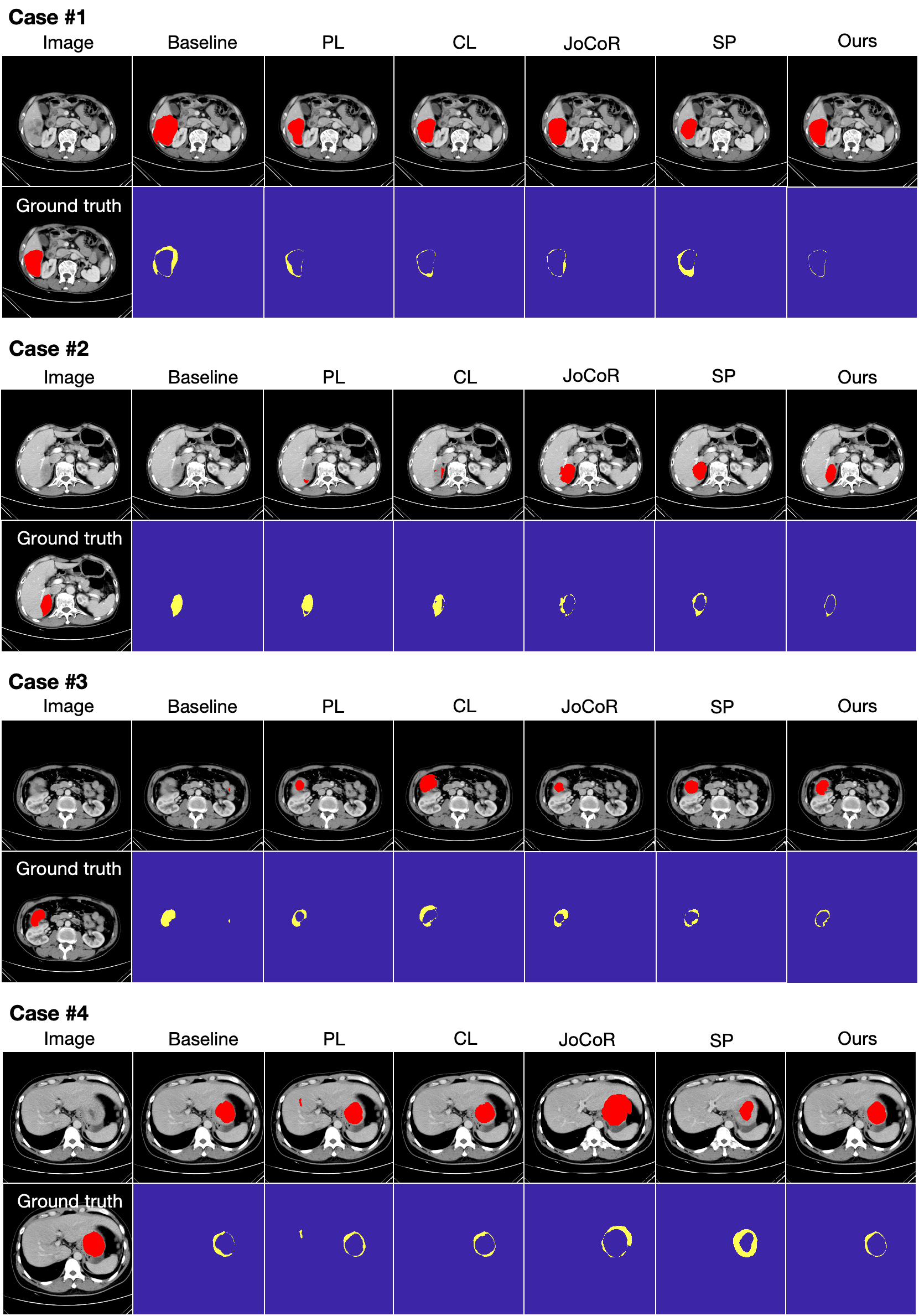}
\caption{Qualitative results of different methods on the HCC dataset (B-model, $\alpha_1=70\%$, $\beta_1=10$), where liver tumor regions are highlighted in red, and discrepancies between network-produced results and ground truths are highlighted in yellow.
}
\label{fig_results}
\end{figure}

\begin{table}[p]
\scriptsize
\renewcommand\arraystretch{1}
\renewcommand\tabcolsep{6pt}
\centering
\caption{Quantitative comparison results in DSC (\%) on the HCC dataset.The Baseline method, trained on the noise-free training set, establishes the upper bound for performance. The B-model is the best checkpoint selected based on the validation set and the L-model is the last checkpoint obtained at the $500^{th}$ epoch. \textcolor{red}{Red} numbers indicate the best results, while \textcolor[rgb]{0.161,0.576,0.835}{blue} numbers indicate the second-best results.}
\vspace{15pt}
\label{tb_results}
\begin{tabular}{ccccc} 
\hline\hline
\multicolumn{2}{c}{\multirow{2}{*}{Noise level}} & \multirow{2}{*}{Method} & \multicolumn{2}{c}{Liver tumor}                  \\ 
\cline{4-5}
\multicolumn{2}{c}{}                             &                         & B-model                & L-model                 \\ 
\hline
\multicolumn{2}{c}{Noise-free (upper bound)}     & Baseline                & 71.45                  & 70.81                   \\ 
\cline{1-5}

\multirow{12}{*}{$\alpha_1=30\%$} & \multirow{6}{*}{$\beta_1=5$}         & Baseline & 70.15 & 67.90  \\
 &  & PL & 70.36 & 64.57  \\
 &  & CL & 70.51 & 63.10 \\
 &  & JoCoR & \textcolor[rgb]{0.161,0.576,0.835}{70.84}  &   68.51  \\
 &  & SP &  70.63 &   \textcolor[rgb]{0.161,0.576,0.835}{68.92}  \\
 &  & Ours & \textcolor{red}{70.85} & \textcolor{red}{69.52}  \\ 
\cline{2-5}

 & \multirow{6}{*}{$\beta_1=10$}   
 & Baseline & 62.36 & 57.72 \\
 &  & PL & 64.68 & 58.94 \\
 &  & CL & 65.56 & 54.53 \\
 &  & JoCoR & \textcolor[rgb]{0.161,0.576,0.835}{69.11}  &   \textcolor[rgb]{0.161,0.576,0.835}{63.32}  \\
 &  & SP & 68.94  &  61.00   \\
 &  & Ours & \textcolor{red}{69.55} & \textcolor{red}{67.76}  \\ 
\hline

\multirow{12}{*}{$\alpha_1=50\%$} & \multirow{6}{*}{$\beta_1=5$}         
 & Baseline  & 66.89  & 64.27  \\
 &  & PL & 67.65 & 65.13 \\
 &  & CL & 68.25  & 64.73 \\
 &  & JoCoR & \textcolor[rgb]{0.161,0.576,0.835}{68.60}  &   65.35  \\
 &  & SP &  68.35  &  \textcolor[rgb]{0.161,0.576,0.835}{65.93}   \\
 &  & Ours & \textcolor{red}{69.07} & \textcolor{red}{66.54}  \\ 
\cline{2-5}

 & \multirow{6}{*}{$\beta_1=10$}  & Baseline & 62.81 & 58.62 \\
 &  & PL  & 62.94  & 59.03 \\
 &  & CL  & 60.88 & 52.11 \\
 &  & JoCoR &  \textcolor[rgb]{0.161,0.576,0.835}{67.90} &  \textcolor[rgb]{0.161,0.576,0.835}{59.81}   \\
 &  & SP & 66.44  &  59.26   \\
 &  & Ours & \textcolor{red}{68.29} & \textcolor{red}{64.31}  \\ 
\hline

\multirow{12}{*}{$\alpha_1=70\%$} & \multirow{6}{*}{$\beta_1=5$}         
 & Baseline  & 65.51 & \textcolor[rgb]{0.161,0.576,0.835}{64.06} \\
 &  & PL & 67.02 & 63.42 \\
 &  & CL & 66.87 & 62.78 \\
 &  & JoCoR &  67.07 &  63.55   \\
 &  & SP &  \textcolor[rgb]{0.161,0.576,0.835}{67.96} &  62.99   \\
 &  & Ours & \textcolor{red}{68.73} & \textcolor{red}{64.38}  \\ 
\cline{2-5}

 & \multirow{6}{*}{$\beta_1=10$} & Baseline & 59.71 & 51.19 \\
 &  & PL & 60.21 & \textcolor[rgb]{0.161,0.576,0.835}{52.50} \\
 &  & CL & 62.43 & 50.16 \\
 &  & JoCoR &  \textcolor[rgb]{0.161,0.576,0.835}{63.38} &  48.77   \\
 &  & SP &  60.73 &   51.79  \\
 &  & Ours & \textcolor{red}{67.02} & \textcolor{red}{63.55}  \\
\hline\hline
\end{tabular}
\end{table}

\subsection{Results}
\label{sec_res}
In this section, we analyze the comparative experimental results between our method and competing methods on the HCC dataset, both qualitatively and quantitatively. First, we analyze the qualitative results from the visual perspective. Fig.~\ref{fig_results} shows several examples of the segmentation results (produced by B-model with $\alpha_1=70\%$, $\beta_1=10$), where difference maps highlight the discrepancies between network-produced results and ground truths. It is evident that the Baseline network is significantly impacted by noisy samples, leading to obvious false negatives (e.g., Case \#2 and Case \#3). While the four competing methods exhibit some noise-resistance ability, there still remains a notable gap between their segmentation results and the ground truths. The issue with PL is that it addresses only image-level label noise by reducing the influence of noisy samples but fails to account for disturbances from local noisy regions within each sample. In contrast, the three pixel-level methods (CL, JoCoR, SP) identify pixel-level noise within all samples indiscriminately and rectify the training process. These methods may either introduce extra noise into clean labels or lead to decreased network efficiency and stability. Particularly, CL performs label cleansing in a one-shot manner, meaning the generated pseudo-labels are fixed and cannot be further refined as training progresses. In contrast, our method demonstrates superior performance in noise reduction and achieves optimal segmentation results. By combining the advantages of both image-level and pixel-level methods, our approach preserves clean samples while effectively cleansing noisy ones during training. Furthermore, our method is an iterative method that updates pseudo-labels to reduce noise throughout the training process. As training progresses, these pseudo-labels are continuously refined, enabling the network to better learn robust target features.

Further, we present quantitative results for different methods in Table~\ref{tb_results}. The Baseline method, trained on the noise-free training set, provides the upper bound for performance. The B-model, representing the best checkpoint selected using the validation set, reflects the actual segmentation performance of the network. The L-model, corresponding to the last checkpoint obtained at the $500^{th}$ epoch, indicates network stability and sustained noise-resistance ability during training. From Table~\ref{tb_results}, it is evident that PL demonstrates robustness to label noise in the B-model and shows some sustained noise resistance in the L-model. CL exhibits noise resistance in the B-model but shows poor stability and sustained noise robustness (as measured by the L-model). As the level of noise increases, (with larger $\alpha_1$ and $\beta_1$), both methods experience a more significant decline in stability and noise resistance. JoCoR and SP show improvements in stability (as reflected in the L-model), but the results are still not ideal. In contrast, our method achieves the best results in both the B-model and L-model. More importantly, as the noise level increases, our method shows increasingly effective noise-resistance performance. Based on both quantitative and qualitative analyses, it can be concluded that our method outperforms competing methods in terms of noise-resistance capability and stability.

\begin{figure}[p]
\centering
\includegraphics[width=\textwidth]{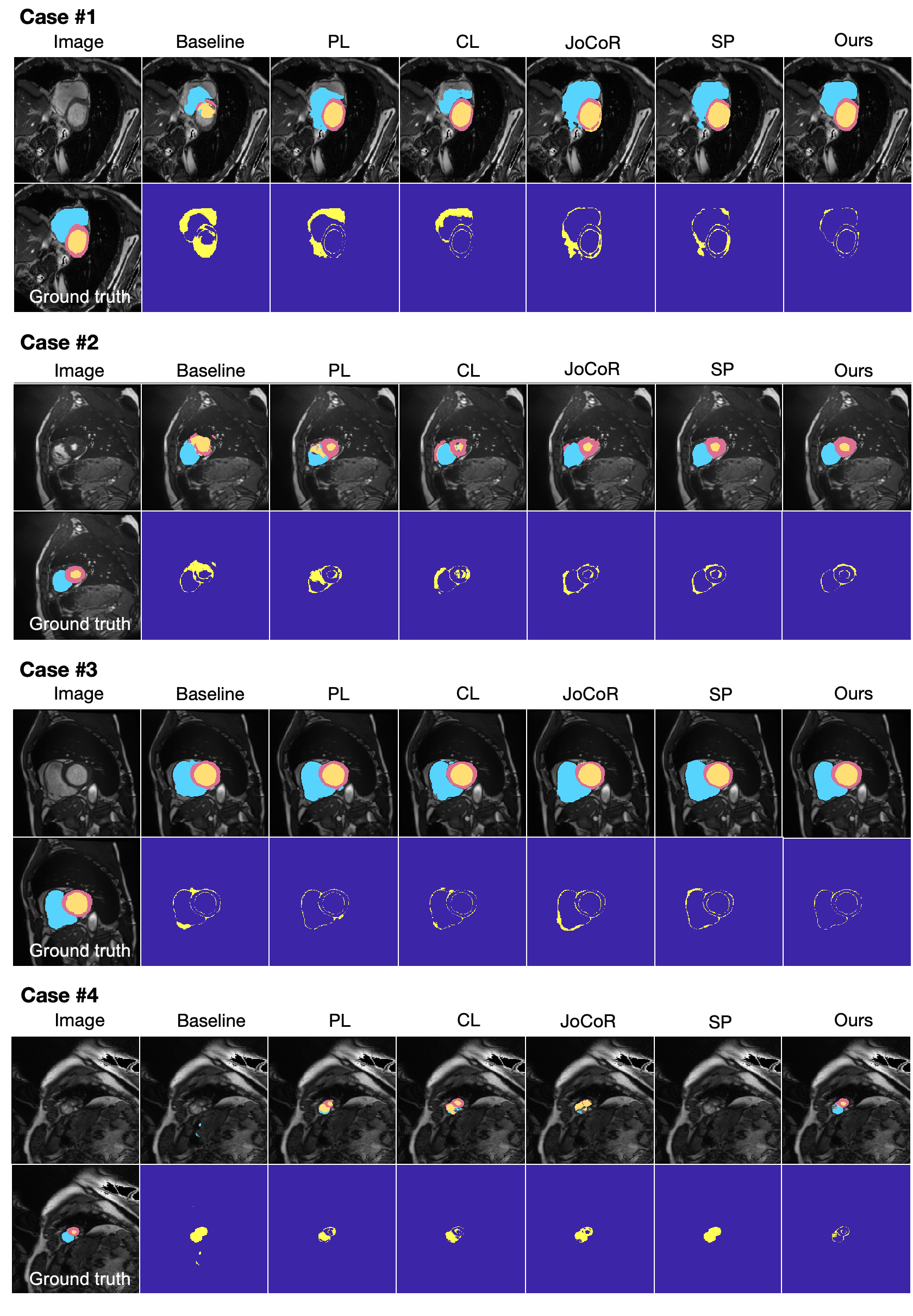}
\caption{Qualitative results of different methods on the ACDC dataset (B-model, $\alpha_2=70\%$, $\beta_2=10$), where right ventricle (RV) regions are highlighted in \textcolor[rgb]{0.34, 0.83, 0.996}{blue}, myocardium (Myo) regions in \textcolor[rgb]{0.863, 0.439, 0.576}{red}, left ventricle (LV) regions in \textcolor[rgb]{1, 0.871, 0.459}{orange}, and discrepancies between network-produced results and ground truths are highlighted in yellow.
}
\label{fig_acdc_results}
\end{figure}

\begin{table*}
\renewcommand\arraystretch{1}
\renewcommand\tabcolsep{1pt}
\scriptsize
\centering
\caption{Generalization validation on the ACDC dataset (measured in DSC (\%)). The Baseline method, trained on the noise-free training set, provides the upper bound for performance. The B-model is the best checkpoint selected using the validation set and the L-model is the last checkpoint obtained in the $500^{th}$ epoch. \textcolor{red}{Red} numbers indicate the best results, while \textcolor[rgb]{0.161,0.576,0.835}{blue} numbers indicate the second-best results.}
\vspace{15pt}
\label{tb_acdc}
\begin{tabular}{ccccccccccc} 
\hline\hline
\multicolumn{2}{c}{\multirow{2}{*}{Noise level}} & \multicolumn{1}{c}{\multirow{2}{*}{Method}} & \multicolumn{2}{c}{RV}                                  & \multicolumn{2}{c}{Myo}                                 & \multicolumn{2}{c}{LV}                              & \multicolumn{2}{c}{Average}                                                            \\ 
\cline{4-11}
\multicolumn{2}{c}{}                             & \multicolumn{1}{c}{}                        & \multicolumn{1}{c}{B-model} & \multicolumn{1}{c}{L-model} & \multicolumn{1}{c}{B-model} & \multicolumn{1}{c}{L-model} & \multicolumn{1}{c}{B-model} & \multicolumn{1}{c}{L-model} & \multicolumn{1}{c}{B-model}               & \multicolumn{1}{c}{L-model}                \\ 
\hline

    \multicolumn{2}{c}{Noise-free}     & Baseline & 76.60 & 72.89  & 78.85 & 77.30 & 83.59 & 81.19 & 79.68 & 77.12 \\ 
    \hline

    \multirow{12}{*}{$\alpha_2=30\%$} & \multirow{6}{*}{$\beta_2=5$}
        
    & Baseline & 76.08 & 59.58 & 75.75 & 58.99 & 81.57 & 65.54 & 77.80 & 61.37 \\
    &   & PL & 73.96 & 59.73 & 75.88 & \textcolor[rgb]{0.161,0.576,0.835}{72.63} & 80.67 & 76.93 & 76.83 & 69.76 \\
    &   & CL & \textcolor[rgb]{0.161,0.576,0.835}{78.89} & 59.37 & 76.09 & 66.89 & \textcolor[rgb]{0.161,0.576,0.835}{82.15} & 72.69 & \textcolor[rgb]{0.161,0.576,0.835}{79.04} & 66.31 \\
    &   & JoCoR &  75.50 &  62.88 &  76.63 &  67.36 &  \textcolor[rgb]{0.161,0.576,0.835}{82.21} & 74.23  &  78.11 & 68.16  \\
    &   & SP & 75.91  & \textcolor[rgb]{0.161,0.576,0.835}{68.07} &  \textcolor[rgb]{0.161,0.576,0.835}{77.09} & 70.78  &  81.63 & \textcolor[rgb]{0.161,0.576,0.835}{78.46}  &  78.21 & \textcolor[rgb]{0.161,0.576,0.835}{72.44}  \\
    &   & Ours &  \textcolor{red}{80.32} & \textcolor{red}{71.66} & \textcolor{red}{78.20} & \textcolor{red}{76.47} & \textcolor{red}{82.47} & \textcolor{red}{80.09} & \textcolor{red}{80.33} & \textcolor{red}{76.07} \\ 
        
    \cline{2-11}

    & \multirow{6}{*}{$\beta_2=10$}

    & Baseline & 74.90 & 53.28 & 77.32 & 60.99 & 81.75 & 67.17 & 77.99 & 60.48 \\
    &   & PL & 75.32 & 58.99 & 75.05 & \textcolor[rgb]{0.161,0.576,0.835}{73.91} & 80.67 & \textcolor[rgb]{0.161,0.576,0.835}{79.90} & 77.01 & \textcolor[rgb]{0.161,0.576,0.835}{70.93} \\
    &   & CL & 76.05 & 58.73 & 77.37 & 66.70 & \textcolor[rgb]{0.161,0.576,0.835}{82.17} & 71.60 & 78.53 & 65.68 \\
    &   & JoCoR & \textcolor[rgb]{0.161,0.576,0.835}{76.87} & 63.57 & 77.26 &  66.81 & 82.13 & 73.87 &  \textcolor[rgb]{0.161,0.576,0.835}{78.75} &  68.08 \\
    &   & SP &  75.30 &  \textcolor[rgb]{0.161,0.576,0.835}{61.64} & \textcolor[rgb]{0.161,0.576,0.835}{77.39}  &  69.30 & 81.42  & 75.57  & 78.04  & 68.83  \\
    &   & Ours & \textcolor{red}{77.87} & \textcolor{red}{69.46} & \textcolor{red}{78.10} & \textcolor{red}{77.25} & \textcolor{red}{84.06} & \textcolor{red}{82.20} & \textcolor{red}{80.01} & \textcolor{red}{76.30} \\ 
    
    \hline

    \multirow{12}{*}{$\alpha_2=50\%$} & \multirow{6}{*}{$\beta_2=5$}  
    
    & Baseline & 71.11 & 41.67 & 72.27 & 42.64 & 78.14 & 51.10 & 73.84 & 45.14 \\
    &   & PL & 69.03 & 61.58 & 74.03 & 63.92 & 63.98 & 61.58 & 69.01 & 62.36 \\
    &   & CL & 70.96 & 55.11 & 73.79 & 52.01 & 79.78 & 58.67 & 74.84 & 55.26 \\
    &   & JoCoR &  71.22 &  56.83 &  73.69 &  56.79 &  \textcolor[rgb]{0.161,0.576,0.835}{80.13} & 64.82  &  75.01 &  59.48 \\
    &   & SP &  \textcolor[rgb]{0.161,0.576,0.835}{71.28} & \textcolor[rgb]{0.161,0.576,0.835}{63.70}  &  \textcolor[rgb]{0.161,0.576,0.835}{74.65} &  \textcolor[rgb]{0.161,0.576,0.835}{64.70} &  79.19 &  \textcolor[rgb]{0.161,0.576,0.835}{75.25} &  \textcolor[rgb]{0.161,0.576,0.835}{75.04} &  \textcolor[rgb]{0.161,0.576,0.835}{67.81} \\
    &   & Ours & \textcolor{red}{73.05} & \textcolor{red}{66.95} & \textcolor{red}{76.06} & \textcolor{red}{74.86} & \textcolor{red}{80.14} & \textcolor{red}{79.27} & \textcolor{red}{76.41} & \textcolor{red}{73.69} \\ 
    
    \cline{2-11}
    
    & \multirow{6}{*}{$\beta_2=10$}     
    
    & Baseline & \textcolor[rgb]{0.161,0.576,0.835}{72.94} & 42.05 & 74.46 & 47.16 & 80.97 & 55.66 & \textcolor[rgb]{0.161,0.576,0.835}{76.12} & 48.29 \\
    &   & PL & 70.54 & \textcolor[rgb]{0.161,0.576,0.835}{66.16} & 71.58 & \textcolor[rgb]{0.161,0.576,0.835}{66.86} & 77.72 & 72.46 & 73.28 & \textcolor[rgb]{0.161,0.576,0.835}{68.49} \\
    &   & CL & 71.88 & 61.61 & \textcolor[rgb]{0.161,0.576,0.835}{74.98} & 53.79 & \textcolor[rgb]{0.161,0.576,0.835}{81.22} & 63.82 & 76.02 & 59.74 \\
    &   & JoCoR &  70.48 &  61.83 & 71.18  & 62.95  &  78.08 &  69.88 & 73.25  &  64.89 \\
    &   & SP &  71.68 &  65.44 & 74.75  & 64.95  & 80.76  & \textcolor[rgb]{0.161,0.576,0.835}{73.75}  & 75.73  &  68.04 \\
    &   & Ours & \textcolor{red}{73.21} & \textcolor{red}{67.12} & \textcolor{red}{75.74} & \textcolor{red}{71.84} & \textcolor{red}{81.33} & \textcolor{red}{75.77} & \textcolor{red}{76.76} & \textcolor{red}{71.57} \\ 
    
    \hline
    
    \multirow{12}{*}{$\alpha_2=70\%$} & \multirow{6}{*}{$\beta_2=5$}     
    
    & Baseline & 61.30 & 23.50 & 58.09 & 16.33 & 69.55 & 27.51 & 62.98 & 22.45 \\
    &   & PL & \textcolor[rgb]{0.161,0.576,0.835}{65.85} & 49.43 & \textcolor[rgb]{0.161,0.576,0.835}{62.87} & 38.03 & 74.10 & 48.55 & \textcolor[rgb]{0.161,0.576,0.835}{67.60} & 45.34 \\
    &   & CL & 62.42 & 46.26 & 57.09 & 37.47 & 66.63 & 49.31 & 62.04 & 44.35 \\
    &   & JoCoR &  62.01 &  38.70 & 61.11  &  32.40 &  \textcolor[rgb]{0.161,0.576,0.835}{75.80} & 38.83  &  66.31 &  36.64 \\
    &   & SP &  65.80 &  \textcolor[rgb]{0.161,0.576,0.835}{56.59} & 62.76  & \textcolor[rgb]{0.161,0.576,0.835}{47.91}  &  73.60 & \textcolor[rgb]{0.161,0.576,0.835}{62.09}  &  67.39 & \textcolor[rgb]{0.161,0.576,0.835}{55.53}  \\
    &   & Ours & \textcolor{red}{77.70} & \textcolor{red}{64.52} & \textcolor{red}{75.65} & \textcolor{red}{68.46} & \textcolor{red}{80.64} & \textcolor{red}{71.73} & \textcolor{red}{78.00} & \textcolor{red}{68.24} \\ 
    
    \cline{2-11}

    & \multirow{6}{*}{$\beta_2=10$}         
    & Baseline & 53.11 & 25.89 & 53.57 & 24.30 & 60.16 & 31.45 & 55.61 & 27.21 \\
    &   & PL & \textcolor[rgb]{0.161,0.576,0.835}{66.26} & 40.66 & \textcolor[rgb]{0.161,0.576,0.835}{72.68} & 40.31 & \textcolor[rgb]{0.161,0.576,0.835}{76.96} & 40.70 & \textcolor[rgb]{0.161,0.576,0.835}{71.96} & 40.56 \\
    &   & CL & 54.94 & 35.53 & 60.12 & 29.42 & 67.32 & 41.60 & 60.79 & 35.52 \\
    &   & JoCoR &  63.36 &  28.20 & 63.09  &  31.58 &  68.72 &  38.60  &  65.05 &  32.79 \\
    &   & SP &  64.32 &  \textcolor[rgb]{0.161,0.576,0.835}{55.39} & 58.69  &  \textcolor[rgb]{0.161,0.576,0.835}{50.43} &  65.54 & \textcolor[rgb]{0.161,0.576,0.835}{61.99}  & 62.58  &  \textcolor[rgb]{0.161,0.576,0.835}{55.94} \\
    &   & Ours & \textcolor{red}{72.45} & \textcolor{red}{69.89} & \textcolor{red}{72.84} & \textcolor{red}{66.84} & \textcolor{red}{80.12} & \textcolor{red}{74.72} & \textcolor{red}{75.14} & \textcolor{red}{70.48} \\ 
    
\hline\hline
\end{tabular}
\end{table*}

\begin{figure}[t]
\small
\centering
\includegraphics[width=0.8\textwidth]{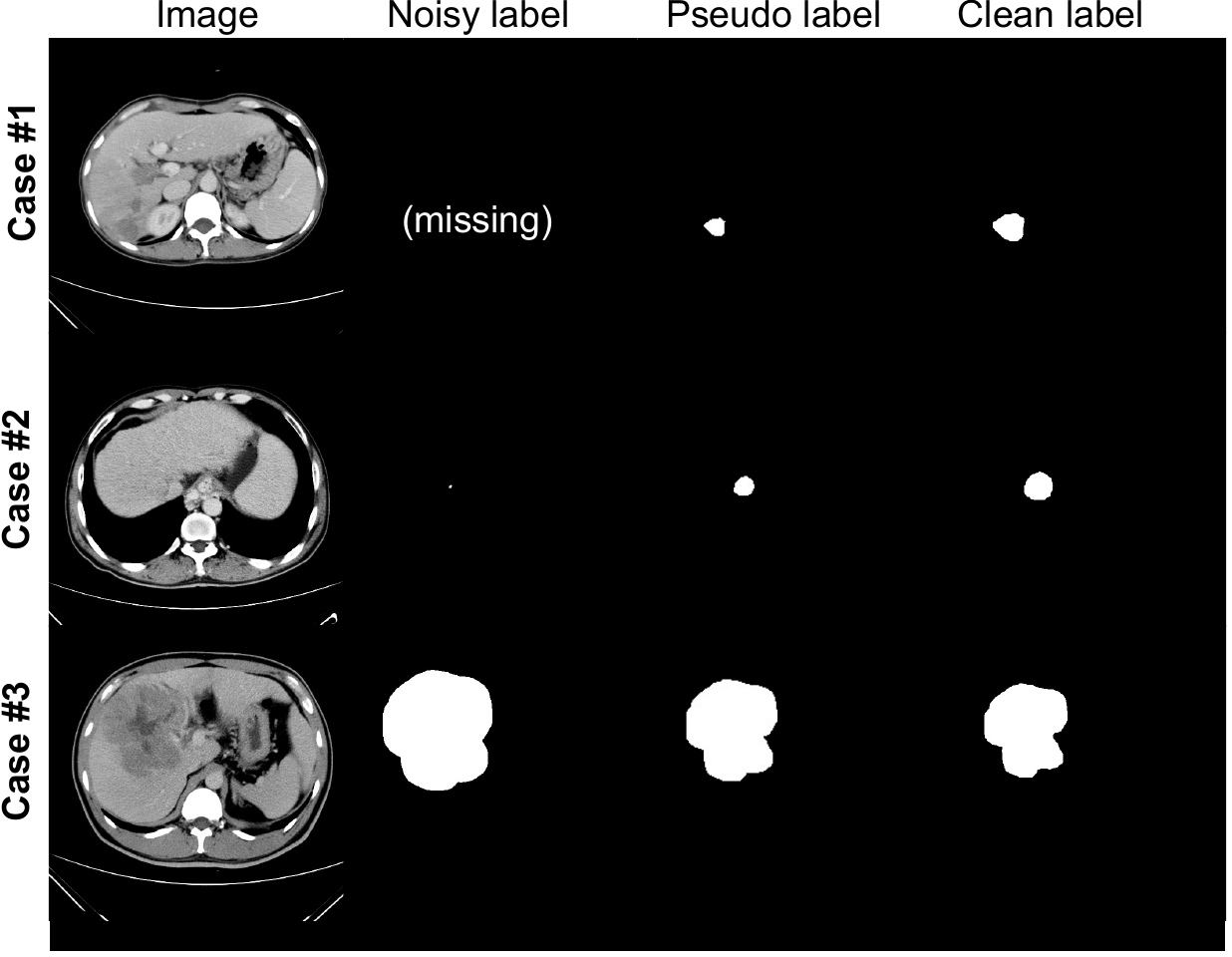}
\caption{Examples of pseudo labels produced by LCM.
}
\label{fig_pseudo_label}
\end{figure}

\begin{table}[t]
\small
\renewcommand\arraystretch{1}
\renewcommand\tabcolsep{7pt}
\caption{Ablation study on the HCC dataset (in DSC(\%)).}
\centering
\begin{tabular}{lcc} 
\hline\hline
Components          & B-model & L-model  \\ 
\hline
Baseline            & 59.71         & 51.19          \\ 
Baseline+LCM        & 61.07         & 53.72          \\ 
Baseline+LCM (pred) & 60.08         & 52.42          \\ 
Baseline+LCM (avg)  & 60.13         & 52.36          \\ 
\textbf{Baseline+LCM+LFM}   & \textbf{67.02}         & \textbf{63.55}          \\
\hline\hline
\end{tabular}
\label{tb_ablation}
\end{table}

\subsection{Generalization Validation}
To assess the generalization ability of the proposed method across different datasets and segmentation tasks, we conduct additional validation experiments on the ACDC dataset. Specifically, we focus on segmenting three different organs—left ventricle (LV), right ventricle (RV), and myocardium (Myo)—from MRI images. Fig.~\ref{fig_acdc_results} provides visual examples of segmentation results produced by different methods (produced by B-model with $\alpha_2=70\%$, $\beta_2=10$). It is shown that our method produces results that are most similar to the ground truths. Table~\ref{tb_acdc} presents the quantitative results, including the analysis of both the B-model and L-model for each method. The results demonstrate that the proposed method outperforms competing methods across the three-organ segmentation tasks. Besides, the improvement of the stability and sustained noise-resistance ability of the network is particularly evident compared with other four methods. These findings on the ACDC dataset demonstrate the generalization capability of our method, implying its noise-resistance ability across different datasets and segmentation tasks.

\subsection{Ablation Study}
To verify the effectiveness of each component of the proposed method, we conduct an ablation study on the HCC dataset (with $\alpha_1=70\%$, $\beta_1=10$). We start from the baseline network (the segmentation backbone without any noise-cleansing strategy) and incrementally add the proposed components. Table.~\ref{tb_ablation} summarizes the results of ablation study.
It is observed that the B-model of the baseline achieves the segmentation performance of 59.71\% (in DSC) on the test set. However, as training proceeds, the network gradually overfits to noisy samples, the segmentation performance eventually falls back to 51.19\% (L-model).

\textbf{Effectiveness of LCM:} We first add LCM to the baseline to assess its effectiveness, whose specific structure is described in Section~\ref{sec_lcm}. After the warm-up training, LCM indiscriminately corrects all samples every 5 epochs without preserving clean samples. Fig.~\ref{fig_pseudo_label}  illustrates several examples of cleansed labels produced by the LCM, demonstrating its effectiveness in correcting both missing labels (Case \#1) and boundary noise (Cases \#2 and \#3). Table~\ref{tb_ablation} provides the corresponding quantitative results, which shows that the LCM improves the segmentation performance of the B-model by +1.36\% (in DSC) and also enhances the stability of the baseline, resulting in a +2.53\% (in DSC) improvement in the performance of the L-model.

Meanwhile, we compare the pseudo-labeling method used in LCM with two other pseudo-labeling methods: \textbf{1)} LCM (pred), which directly employs network predictions as pseudo labels; \textbf{2)} LCM (avg), which yields pseudo labels based on averaged prototypes (the averaged features of suspected target regions). The experimental results of the comparison are shown in Table~\ref{tb_ablation}. The results indicate that LCM based on representative prototypes produces more accurate pseudo-labels as it generates better representative features of the target regions.

\textbf{Effectiveness of LFM:} Although LCM effectively improves label quality, indiscriminately modifying all labels can corrupt clean and low-noise manual labels, as the generated pseudo-labels still contain some noise. Such noise can accumulate as training progresses, which potentially limits the performance and stability of the network.

To address this issue, we introduce LFM to allow the network to preserve clean samples and correct only noisy samples in each cleansing iteration, thereby reducing the potential noise introduced by pseudo-labeling. LFM achieves this by modeling per-sample loss distributions and classify labels into \emph{Clean} and \emph{Noisy} categories. Incorporating LFM ensures that the network is consistently supervised by low-noise manual labels, which stabilizes the training process and prevents error accumulation. As shown in Table~\ref{tb_ablation}, LFM significantly boosts performance by +5.95\% in the B-model and +9.83\% in the L-model, which demonstrates its effectiveness.

\section{Discussion}
In this paper, we propose a self-cleansing network for medical image segmentation that is highly resistant to label noise. To achieve this, our method starts by filtering out noisy labels that require cleaning while preserving clean labels based on per-sample loss distributions. It then generates pseudo-labels for the noisy labels using representative target prototypes. Finally, both the preserved clean labels and the generated pseudo-labels are used together to supervise the network. The experimental results of our method are detailed in Section~\ref{sec_res}. We validate its effectiveness on the HCC dataset of liver tumors and it shows superior sustained noise-resistance compared to competing methods. Additionally, validation on the publicly available ACDC dataset for automatic cardiac diagnosis demonstrates the generation ability of our method across diverse datasets and segmentation tasks.

The proposed method offers two main advantages. Firstly, it effectively preserves clean and low-noise labels in each self-cleansing iteration. This is achieved by identifying image-level label noise from both manual labels and previously generated pseudo-labels using LFM. This strategy allows the network to cleanse only a subset of samples in each iteration, ensuring that a certain portion of clean labels are preserved and enhancing stability throughout the training process. Secondly, the method employs continuous representative regions as prototypes for updating pseudo-labels through LCM. These prototypes accurately capture local information within target regions, which helps to generate more precise pseudo-labels.

Despite these advantages, our approach has certain limitations. For example, the framework relies on a single segmentation network for label noise cleansing. Once the network begins fitting noisy labels, the LFM based on loss distributions may struggle to accurately assess the noise level. As training progresses, the network might amplify these errors, which is a phenomenon known as confirmation bias~\cite{liu2021s}. This can negatively impact the network's performance and lead to a gradual decline in accuracy during the later stages of training. To address this issue, future work could introduce a co-training approach. Such approach would involve training two segmentation networks simultaneously and allow them to mutually identify and correct label noise, thus avoiding confirmation bias. Each network would update its pseudo-labels based on the other network’s predictions, and both networks would adjust their parameters using the pseudo-labels generated by their counterpart. This co-training mode could help mitigate confirmation bias caused by using a single network, which has the potential to further improve the performance.

\section{Conclusion}
In this paper, we propose a deep self-cleansing network for medical image segmentation to effectively reduce label noise in training datasets. Our network is designed to preserve clean labels while iteratively cleansing noisy labels throughout the training process. We develop a Label Filtering Module based on a gaussian mixture model to distinguish between noisy and clean labels. Additionally, we introduce a Label Cleansing Module that uses pseudo-labeling to correct the identified noisy samples. Experimental evaluations on a clinical CT dataset of liver tumors and a publicly available MRI dataset for automated cardiac diagnosis show that our method consistently outperforms several competing methods. Furthermore, the results demonstrate that our method is robust when dealing with varying levels of label noise.

\bibliographystyle{plain} 
\bibliography{mybib}

\end{document}